\documentclass[runningheads]{llncs}

\usepackage{eccv}

\usepackage{eccvabbrv}

\usepackage{graphicx}
\usepackage{booktabs}

\usepackage[accsupp]{axessibility}  

\usepackage{amsmath}
\usepackage{booktabs}
\usepackage{multirow} 
\usepackage{bbding}
\usepackage{tabularx}
\usepackage{multicol}

\usepackage{hyperref}

\usepackage{orcidlink}

\begin{document}

\title{OBBSeg: Irregular Lesion Segmentation under Oriented Bounding Box Annotations}

\titlerunning{OBBSeg}
\authorrunning{Wei et al.}

\author{
    Jun Wei\inst{1}\orcidlink{0000-0002-1263-5260}\and 
    Xinchang Liu\inst{1}\and 
    Yu Liu\inst{1}\and
    Chuhua Yang\inst{1}\and
    Shuhui Wang\inst{2}\and \\
    Hui Huang\inst{1}\thanks{Corresponding author}\orcidlink{0000-0003-3212-0544}
}

\institute{
    Guangdong Provincial Key Laboratory of Visual Media and Multidimensional Intelligence, CSSE, Shenzhen University, China \and
    Institute of Computing Technology, Chinese Academy of Sciences, China \\
    \email{weijun@szu.edu.cn, hhzhiyan@gmail.com}
}
 
\maketitle

\begin{abstract}
    Pixel-level annotation remains a major bottleneck in medical image segmentation, making weak supervision an attractive yet under-constrained alternative. We propose \textbf{OBBSeg}, an intermediate supervision paradigm guided by \textbf{Oriented Bounding Boxes (OBBs)} that bridges the gap between full and weak supervision. By jointly encoding spatial extent and orientation, OBBs provide compact geometric supervision that better aligns with elongated or anisotropic lesions, reducing the ambiguity of coarse box annotations. To mitigate the inherent rectangular bias of OBBs, we introduce a \textbf{Mask-to-OBB loss}, a differentiable formulation that enforces geometric consistency between predicted masks and OBB regions. Furthermore, we incorporate \textbf{prompt-driven semantic guidance} through two complementary modules—\textbf{PAFE} and \textbf{DBFE}—which enhance foreground representation and suppress background interference. Extensive experiments on \textbf{13 datasets} across \textbf{5 imaging modalities} show that OBBSeg not only outperforms existing weakly supervised methods but also achieves performance comparable to fully supervised approaches, demonstrating its potential for \textbf{efficient and scalable medical image segmentation}. The code is available at \url{https://github.com/StarLxc3/OBBSeg}.
\end{abstract}

\section{Introduction}
\label{sec:intro}

\begin{figure}[!t]
    \centering  
    \includegraphics[width=0.6\columnwidth]{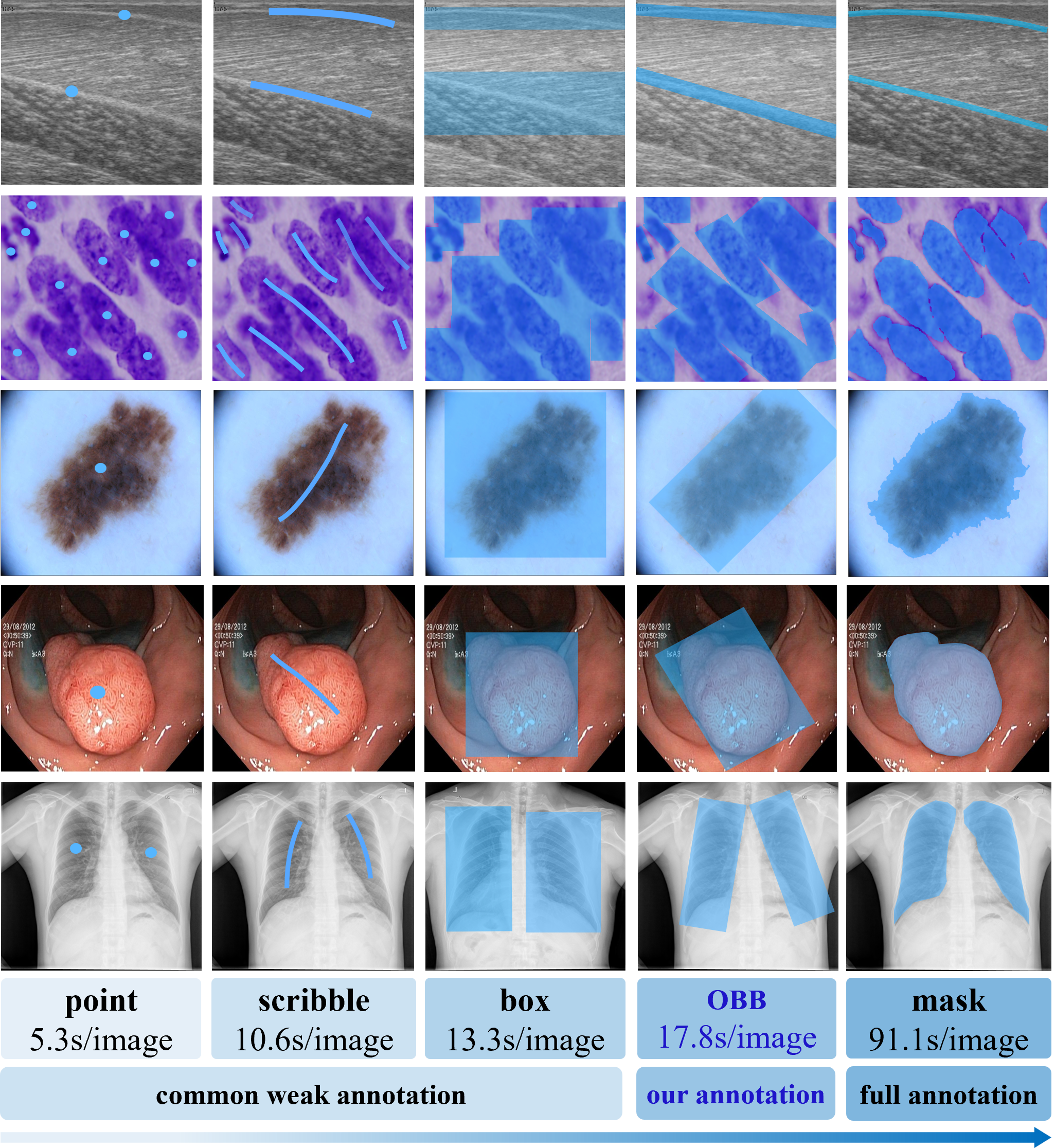}
    \caption{Visualization and annotation cost (seconds per image) comparison among different annotation types ({\it e.g.}, point, scribble, box, OBB, and mask). OBB annotation achieves a favorable balance between labeling efficiency and target coverage, providing geometry-aware and orientation-sensitive supervision with minimal annotation cost.}
    \label{fig:teaser}
\end{figure}

Accurate segmentation of medical images plays a critical role in computer-aided diagnosis and treatment planning. By delineating anatomical structures and pathological regions, segmentation models enable reliable disease screening, quantitative analysis, and therapy assessment.

Despite recent advances, medical image segmentation still faces a fundamental trade-off between \textbf{annotation accuracy and efficiency}. Fully supervised models, such as U-Net~\cite{RN6} and its variants~\cite{RN7}, achieve strong performance with pixel-level annotations, but dense mask labeling is expensive and time-consuming, especially for lesions with irregular or ambiguous boundaries.
To alleviate this burden, weakly supervised segmentation methods employ coarse annotations such as points~\cite{RN45}, scribbles~\cite{RN17}, or bounding boxes~\cite{RN14}. These annotations significantly reduce labeling effort but provide limited geometric constraints. In particular, conventional bounding boxes fail to encode orientation, making them inadequate for \textbf{elongated or anisotropic lesions}.

To bridge the gap between full and weak supervision, we propose an intermediate paradigm, termed \textbf{Oriented Bounding Box Guided Segmentation (OBBSeg)}, combining the efficiency of weak supervision with the geometric fidelity of full supervision.
Fig.~\ref{fig:teaser} illustrates the proposed Oriented Bounding Box (OBB) annotations, which provide a compact, orientation-aware annotation that better captures lesion geometry than conventional bounding boxes.
By jointly encoding spatial extent and orientation, OBBs tightly fit elongated or anisotropic lesions (see row 1), providing stronger geometric constraints and reducing ambiguous regions (see rows 3-5). 
Moreover, they alleviate the overlap between closely packed lesions that commonly occurs with conventional box annotations (see row 2).

Furthermore, OBB annotation remains \textbf{lightweight}, requiring only four corner points per instance. To evaluate its annotation cost, we conducted a human annotation study using our training set, where three independent annotators labeled 1,616 OBBs across 1,450 images. The average annotation time was \textbf{17.8 s per OBB}, only slightly longer than standard bounding boxes (13.3 s per image) and substantially shorter than dense mask annotation (91.1 s per image). These results show that OBB annotation provides richer geometric information with only a modest increase in annotation effort, making it an effective compromise between annotation efficiency and geometric precision.


Although OBB annotations capture target extent and orientation, their rectangular geometry may introduce shape bias during segmentation learning. To address this limitation, we propose the \textbf{Mask-to-OBB (M2O) loss}, a supervision strategy that bridges coarse OBB annotations and fine-grained segmentation masks. M2O projects predicted masks into the OBB space through rotation and projection operations, supervising only spatial extent and orientation while excluding explicit boundary information. This design effectively mitigates rectangular shape bias without iterative refinement. Implemented entirely with matrix operations, M2O is end-to-end \textbf{differentiable, plug-and-play transferable} to other segmentation tasks, and incurs \textbf{no additional inference cost}. As shown in Table~\ref{tab:ablation}, M2O consistently outperforms direct OBB supervision, demonstrating the effectiveness of the proposed supervision strategy.

Although OBB supervision reduces annotation cost and provides geometric priors, it remains limited in modeling complex lesion appearance and cross-domain variation. To address this, we propose a unified OBBSeg framework that integrates OBB-based geometric constraints with prompt-guided semantic learning. Different from existing prompt-based methods that introduce prompts only at the decoder stage, OBBSeg injects prompt information into the encoder to explicitly guide feature learning with semantic priors.

In this design, OBBs provide structural cues of spatial extent and orientation, while prompts progressively refine semantic representations and enhance boundary awareness during feature extraction. To further strengthen weakly supervised learning, we introduce two complementary modules: \textbf{Prompt-assisted Foreground Enhancer (PAFE)}, which highlights lesion regions using prompt-derived masks while suppressing background interference, and the \textbf{Differential-based Foreground Enhancer (DBFE)}, which improves feature discriminability by explicitly enhancing foreground–background contrast.

Overall, OBBSeg jointly leverages geometric constraints, prompt-driven semantics, and differential feature enhancement to improve segmentation accuracy and generalization under weak supervision. Our contributions are fourfold:
\begin{itemize}
    \item[1.] We propose \textbf{OBBSeg}, a new segmentation paradigm that leverages oriented bounding boxes to bridge the gap between weakly and fully supervised learning, providing geometry-aware supervision with low annotation cost.
    
    \item[2.] We introduce the \textbf{M2O loss}, a differentiable and noise-free supervision formulation that removes the rectangular bias of box annotations and enables geometry-consistent learning.
    
    \item[3.] We integrate \textbf{prompt-guided learning} with OBB supervision and design two complementary modules, \textbf{PAFE} and \textbf{DBFE}, to enhance foreground representation and improve segmentation robustness.
    
    \item[4.] Extensive experiments on \textbf{13 medical datasets} across \textbf{5 imaging modalities} demonstrate that OBBSeg outperforms existing weakly supervised methods and achieves performance approaching fully supervised models.
\end{itemize}

\section{Related Work}

\textbf{Medical Image Segmentation.}
Medical image segmentation aims to delineate anatomical and pathological structures from modalities such as X-ray~\cite{RN47}, MRI~\cite{RN44}, and ultrasound~\cite{RN28}. Fully supervised models, including U-Net~\cite{RN6}, U-Net++~\cite{RN7}, and PraNet~\cite{RN8}, have achieved strong performance across various medical imaging tasks. However, these methods rely heavily on pixel-level annotations, which are expensive and time-consuming to obtain. To reduce annotation costs, weakly supervised segmentation methods have been increasingly explored. In this work, we adopt oriented bounding boxes as supervision, which provide geometry-aware annotations that better capture object extent and orientation while maintaining low labeling cost.

\textbf{Weakly Supervised Learning.}
Weakly supervised segmentation aims to reduce annotation effort by using coarse labels such as points~\cite{RN45}, boxes~\cite{RN14}, scribbles~\cite{RN17}, and circles~\cite{RN48}. \cite{RN14} proposed a box-supervised approach with Background-Aware Pooling and Noise-Aware Loss, achieving competitive results on the PASCAL VOC 2012 benchmark~\cite{RN13}. WeakPolyp~\cite{RN9} introduced a projection–backprojection strategy together with Scale Consistency Loss to mitigate box-shape bias, significantly improving polyp segmentation performance. Zhang \emph{et al.}~\cite{RN16} proposed a Class-Driven Scribble Promotion network that integrates scribble annotations with image-level pseudo-labels and uncertainty-aware optimization. Despite these advances, weak supervision often struggles to capture accurate geometric structures due to the coarse nature of the annotations. In contrast, our method introduces the Mask-to-OBB loss to provide geometry-consistent supervision while eliminating the rectangular bias of box annotations.

\textbf{Prompt-based Image Segmentation.}
Prompt-based segmentation leverages auxiliary guidance, such as points, boxes, or textual cues, to improve model adaptability and generalization. Representative models such as SAM~\cite{RN11} and SAM2~\cite{RN12} encode prompts into geometric embeddings to guide segmentation. However, most existing approaches introduce prompts only at the decoder stage, limiting their influence on feature representation. In contrast, our method incorporates prompts into the encoder to enhance foreground representations and suppress background noise, while further utilizing them for supervision, enabling end-to-end prompt-guided segmentation.

\begin{figure*}[htp]
    \centering  
    \includegraphics[width=\textwidth]{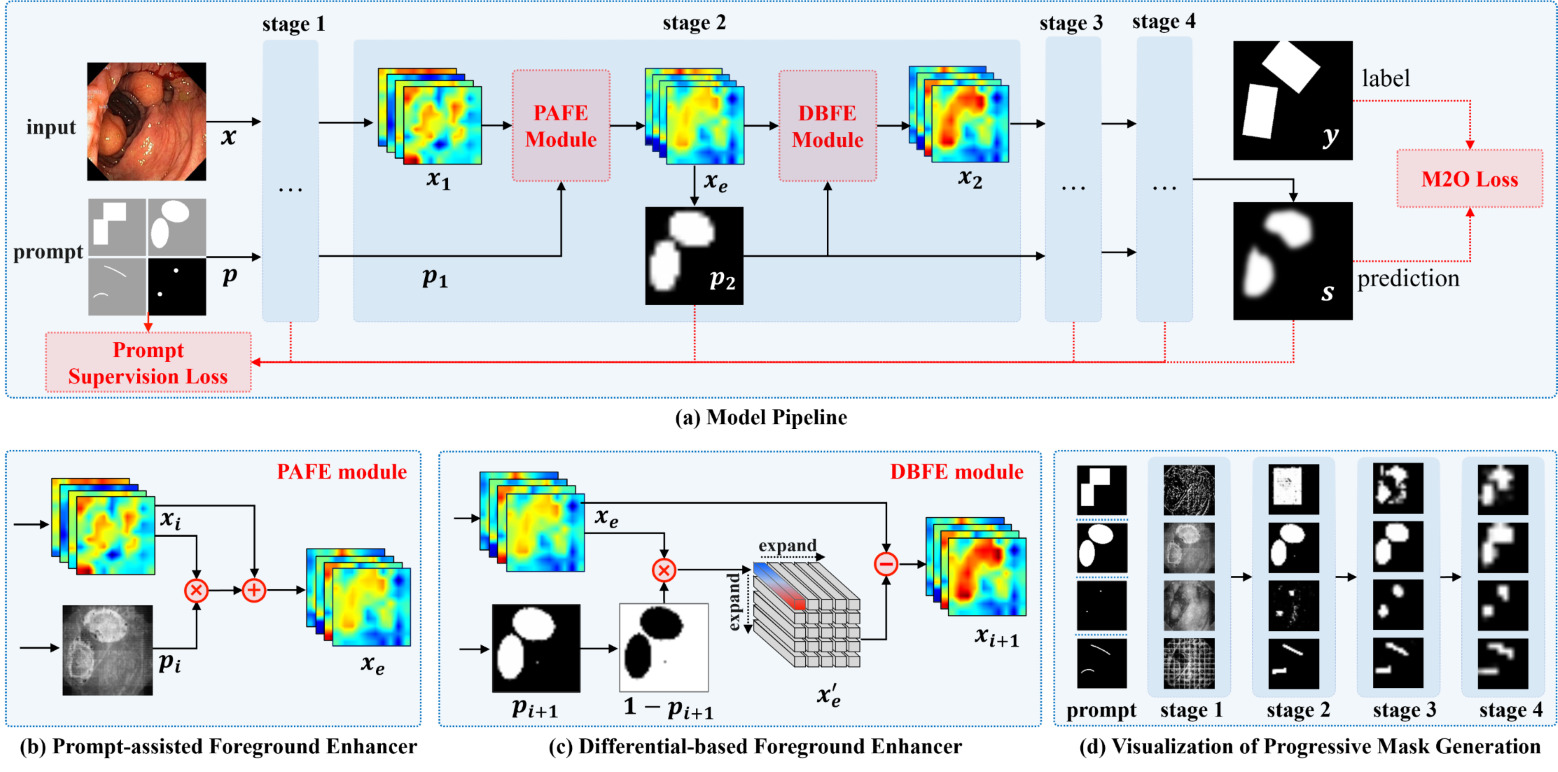}
    \caption{Overview of the proposed OBBSeg framework. (a) Overall architecture. (b) The PAFE module enhances lesion-focused features via prompt guidance. (c) The DBFE module improves foreground--background discrimination. (d) Progressive mask generation from prompt inputs.}
    \label{fig:pipeline}
\end{figure*}

\section{Proposed Methodology}
Fig.~\ref{fig:pipeline}(a) presents the overall architecture of the proposed OBBSeg framework, which leverages prompts as auxiliary cues and oriented bounding boxes as weak supervision.
Built upon a ViT~\cite{dosovitskiy2021vit} backbone, OBBSeg is organized into four stages. In each stage, the PAFE module exploits prompt cues to suppress background noise, while the DBFE module enhances foreground–background contrast to highlight discriminative regions.
At the supervision level, OBBSeg adopts a dual-strategy design. Prompt supervision provides hierarchical guidance to refine coarse prompts, while the proposed M2O loss aligns soft mask predictions with OBB annotations to alleviate box-induced shape bias.
Through progressive feature refinement and noise-robust supervision, OBBSeg produces accurate segmentation masks with minimal annotation cost.

\begin{figure*}[htp]
    \centering
    \includegraphics[width=1\textwidth]{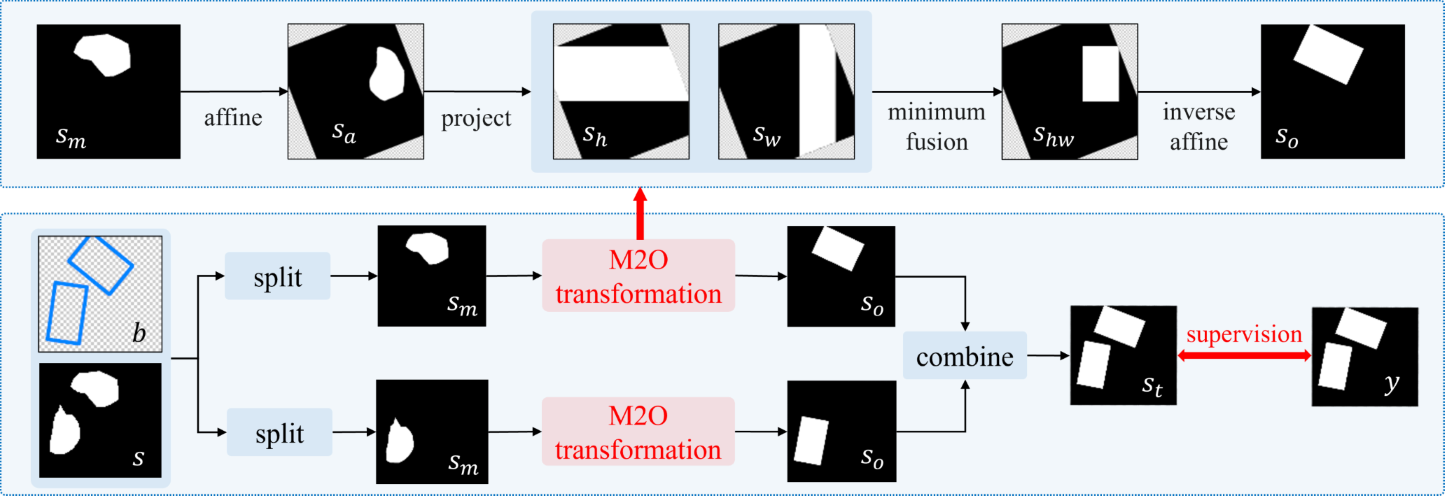}
    \caption{Overview of the proposed Mask-to-OBB (M2O) loss. The predicted mask $s$ is projected into the OBB space to preserve spatial extent and orientation while suppressing fine-grained shape details. The projected representation is aligned with the OBB label $y$, enabling geometry-consistent learning from OBB annotations.}
    \label{fig:M2O}
\end{figure*}

\subsection{Prompt-assisted Foreground Enhancer}
    
Fig.~\ref{fig:pipeline}(b) illustrates the Prompt-assisted Foreground Enhancer (PAFE), designed to suppress background noise under weak supervision. Unlike conventional attention mechanisms that lack explicit guidance, PAFE utilizes prompt cues to guide the network toward lesion-relevant regions, improving semantic consistency and interpretability.
Given an input image $x$, the backbone extracts a hierarchy of feature maps $\{x_i \mid i \in (0,1,2,3)\}$. For each stage $i$, a prompt-derived binary mask $p_i$ is generated and resized to match the spatial resolution of $x_i$.
Foreground-aware filtering is achieved by element-wise multiplication ($\odot$) between $p_i$ and $x_i$. Since prompts (e.g., points or scribbles) may only partially cover the target region, a residual connection adds the original feature map back to preserve information, producing the enhanced feature representation $x_e$:
\begin{equation}
    x_e = x_i \odot p_i + x_i.
    \label{eq:pafe}
\end{equation}
To obtain the prompt mask $p_i$, we directly leverage user-provided prompts (e.g., points or scribbles) as partial supervision signals. Supervision is enforced only on confidently labeled pixels within the prompt regions, providing reliable and noise-free guidance. This partial supervision encourages the model to progressively learn object-aware masks aligned with the target’s location and shape, as illustrated in Fig.~\ref{fig:pipeline}(d).

\subsection{Differential-based Foreground Enhancer}
Medical image segmentation often suffers from high visual similarity between foreground and background, which hinders discriminative feature learning, especially under limited training data. To address this issue, we propose the Differential-based Foreground Enhancer (DBFE), which explicitly amplifies foreground–background contrast to facilitate feature discrimination. The overall architecture is shown in Fig.~\ref{fig:pipeline}(c).

Given the prompt-derived binary mask $p_{i+1}$ highlighting the foreground region, its complement $(1 - p_{i+1})$ represents the background. We first extract background cues by applying the background mask to the input feature $x_e$ via element-wise multiplication followed by average pooling, producing a background representation $x_e^{'}$.
To enhance feature contrast, a differential factor is computed by subtracting $x_e^{'}$ from $x_e$ and applying the foreground mask $p_{i+1}$ through element-wise multiplication ($\odot$), which emphasizes discriminative foreground information. Following a residual design, this differential factor is added back to the original feature to preserve contextual information, yielding the enhanced feature $x_{i+1}$.

\begin{equation}
    \begin{aligned}
        x_e^{'} &= \text{pooling}((1 - p_{i+1}) \odot x_e) \\
        x_{i+1} &= (x_e - x_e^{'}) \odot p_{i+1} + x_e
    \end{aligned}
    \label{eq:dbfe}
\end{equation}

\subsection{Mask-to-OBB Loss}
Direct supervision with OBBs introduces rectangular shape bias, misleading the model toward box-like predictions. To address this, we propose Mask-to-OBB (M2O) loss, which removes shape bias by aligning predictions with OBBs in a geometry-aware way.
Specifically, we extract instance masks, rectify them via affine transformation, and eliminate shape details through axis-wise projection and minimum fusion. The resulting masks retain spatial extent without contour bias, and are inversely transformed back for pixel-wise loss computation. This enables effective learning from coarse OBB annotations without shape-induced errors.

\textbf{Split.}
Given the predicted mask $s$ and a set of OBB masks $\{b_1, b_2, \dots, b_M\}$, we first decompose $s$ into instance-level sub-masks, each corresponding to an individual object. Specifically, for each OBB mask $b_m$, the sub-mask is extracted by element-wise multiplication $s_m = s \odot b_m$.

\textbf{Affine Transformation.}
Given an OBB mask $b_m$, we extract its rotation angle $\theta_m$ and construct the affine transformation matrix $R_m$. This matrix rotates the sub-mask $s_m$ by $\theta_m$ clockwise, resulting in the rectified prediction $s_a$ which is aligned to axis orientation.
\begin{equation}
    s_a =  \text{Warp}(s_m, R_m)
    \label{eq:affine}
\end{equation}

\textbf{Projection.}
We perform max-projection on $s_a$ along the horizontal and vertical axes, yielding the maximum probability vectors: $s_h$ and $s_v$. This step completely eliminates the shape and contour information of $s_a$, retaining only its spatial location and dimension information.
\begin{equation}
    \begin{aligned}
        s_h &= \text{max}(s_a, \text{axis}=1)\in [0,1]^{1 \times W} \\
        s_w &= \text{max}(s_a, \text{axis}=0)\in [0,1]^{H \times 1} \\
    \end{aligned}
    \label{eq:projection}
\end{equation}

\textbf{Minimum Fusion.}
To reconstruct the spatial mask, we back-project the horizontal and vertical projection vectors by replicating $s_h$ across width and $s_w$ across height, yielding two 2D masks $s_h \in [0,1]^{H\times W}$ and $s_w \in [0,1]^{H\times W}$. These masks are then fused via element-wise minimum operation to produce the final aligned prediction $s_{hw}$.
\begin{equation}
    \begin{aligned}
        s_h    &= \text{repeat}(s_h, H, \text{axis}=1) \in [0,1]^{H\times W} \\
        s_w    &= \text{repeat}(s_w, W, \text{axis}=0) \in [0,1]^{H\times W} \\
        s_{hw} &= \min(s_h, s_w) \in [0,1]^{H\times W} \\
    \end{aligned}
    \label{eq:back-projection}
\end{equation}

\textbf{Inverse Affine Transformation.}
We apply the inverse transformation matrix $R_m^{-1}$ to $s_{hw}$, rotating it counterclockwise by $\theta_m$ back to an axis-aligned state. This produces the transformed predicted mask $s_o$.
\begin{equation}
    \begin{aligned}
        s_o &= \text{Warp}(s_{hw}, R_m^{-1})
    \end{aligned}
    \label{eq:inverse:affine}
\end{equation}

\textbf{Combination.}
For each image, we repeat the above transformation process for all OBBs, generating $M$ rectified prediction masks ${s_o^1, s_o^2, \dots, s_o^m}$. We then aggregate them via pixel-wise maximum fusion to obtain the final transformed mask $s_t$:
\begin{equation}
    s_t = \max(s_o^1, s_o^2, \dots, s_o^M) \in [0,1]^{H\times W}
    \label{eq:combination}
\end{equation} 

\textbf{Supervision.}
The transformed prediction $s_t$ produced by the M2O transformation enables direct computation of Binary Cross-Entropy loss $\mathcal{L}_{\text{BCE}}$ and Dice loss $\mathcal{L}_{\text{Dice}}$ against the box-shaped OBB ground truth $y$. The combined loss $\mathcal{L}_{\text{M2O}}$ is defined as Eq.~\ref{eq:M2O}. Definitions of $\mathcal{L}_{\text{BCE}}$ and $\mathcal{L}_{\text{Dice}}$ are provided in the supplementary material.
\begin{equation}
    \mathcal{L}_{\text{M2O}} = \mathcal{L}_{\text{BCE}}(s_t, y) + \mathcal{L}_{\text{Dice}}(s_t, y)
    \label{eq:M2O}
\end{equation}

\subsection{Prompt Supervision Loss}
To compensate for the limited supervision from OBB annotations and better exploit prompt masks, we introduce the Prompt Supervision Loss $\mathcal{L}_{\text{PS}}$. This loss serves two purposes: (1) guiding early-stage feature learning to refine intermediate predictions $p_i$, and (2) strengthening supervision to complement OBB-based learning.
Specifically, $\mathcal{L}_{\text{PS}}$ enforces hierarchical consistency between the intermediate predictions and the final prediction with respect to the prompt mask:
\begin{equation}
\mathcal{L}_\text{PS} = \mathcal{L}(s, p) + \sum_{i=1}^4 \mathcal{L}(p_i, p).
\end{equation}
To support different prompt types (\eg, box, point, scribble, circle), we adopt adaptive supervision functions $\mathcal{L}$ tailored to each prompt. This results in $\mathcal{L}_{\text{PS}}^{\text{box}}$, $\mathcal{L}_{\text{PS}}^{\text{scribble}}$, $\mathcal{L}_{\text{PS}}^{\text{point}}$, and $\mathcal{L}_{\text{PS}}^{\text{circle}}$. Detailed implementations for each prompt type are provided in the supplementary material.

\subsection{Total Loss}
Following WeakPolyp, we adopt the Scale Consistency Loss $\mathcal{L}_{\text{SC}}$ to alleviate the supervision deficiency of OBB annotations by enforcing prediction consistency across different input scales. Combined with the proposed $\mathcal{L}_{\text{M2O}}$ and $\mathcal{L}_{\text{PS}}$, the overall loss function is defined as Eq.~\ref{eq:loss:total}, where $\lambda$ is a balancing hyper-parameter. Detailed analysis of $\lambda$ is provided in the supplementary material.
\begin{equation}
    \mathcal{L}_{\text{Total}} = \mathcal{L}_{\text{M2O}} + \lambda \mathcal{L}_{\text{PS}} + \mathcal{L}_{\text{SC}}
    \label{eq:loss:total}
\end{equation}


\section{Experimental Results}
\textbf{Datasets.}
We evaluate OBBSeg on \textbf{13 public datasets across five imaging modalities}: colonoscopy (Kvasir~\cite{RN25}, ClinicDB~\cite{RN22}, ColonDB~\cite{RN23}, ETIS~\cite{RN24}, Endo~\cite{RN26}, SUN-SEG~\cite{RN18}), dermoscopy (ISIC-2017~\cite{RN19}, ISIC-2018~\cite{RN20,RN21}), MRI (NCI-ISBI~\cite{RN54}), CT (Synapse~\cite{RN55}), and ultrasound (TN3K~\cite{RN28}, DDTI~\cite{RN29}, BUSI~\cite{RN27}). Detailed descriptions are provided in the supplementary material.

\begin{table}[htp]
    \centering
    \caption{Comparison of OBBSeg with fully supervised models on polyp datasets. Dice (\%) is reported. "-" indicates missing results. \textcolor{red}{Avg.} denotes the weighted average across datasets and serves as the primary indicator. Best results are in bold.}    
    \resizebox{\textwidth}{!}{
    \begin{tabular}{l l c | c c c c c c | c c c}
        \toprule
        \multirow{2}{*}{\textbf{Methods}} & \multirow{2}{*}{\textbf{Year}} & \multirow{2}{*}{\textbf{Supervision}}  & \multicolumn{6}{c|}{\textbf{Five Standard Polyp Datasets (FSPD)}} & \multicolumn{3}{c}{\textbf{SUN-SEG}}\\
        \cmidrule(lr){4-12}
        & & & \textbf{ClinicDB} & \textbf{ColonDB} & \textbf{ETIS} & \textbf{Kvasir} & \textbf{Endo} & \textcolor{red}{\textbf{Avg.}}  & \textbf{Easy} & \textbf{Hard} & \textcolor{red}{\textbf{Avg.}}\\ 
        \midrule
        U-Net~\cite{RN6}      & 2015\tiny{MICCAI} & full & 82.3 & 51.2 & 39.8 & 81.8 & 71.0 & \textcolor{red}{56.1} & -    & -    & -\\
        U-Net++~\cite{RN7}    & 2018\tiny{DLMIA}  & full & 79.4 & 48.3 & 40.1 & 82.1 & 70.7 & \textcolor{red}{54.6} & -    & -    & -\\
        PraNet~\cite{RN8}     & 2020\tiny{MICCAI} & full & 89.9 & 70.9 & 62.8 & 89.8 & 87.1 & \textcolor{red}{74.0} & 68.9 & 66.0 & \textcolor{red}{67.7} \\
        SANet~\cite{RN35}      & 2021\tiny{MICCAI} & full & 91.6 & 75.3 & 75.0 & 90.4 & 88.8 & \textcolor{red}{79.4} & 69.3 & 69.4 & \textcolor{red}{69.3} \\
        MEGANet~\cite{RN39}    & 2024\tiny{WACV}   & full & 93.8 & 79.3 & 73.9 & 91.3 & 89.9 & \textcolor{red}{81.4} & 61.2  & 69.3 & \textcolor{red}{64.6} \\
        CASCADE~\cite{RN40}    & 2023\tiny{WACV}   & full & 94.3 & 82.5 & 80.1 & 92.6 & 90.5 & \textcolor{red}{84.7} & -    & - & -\\
        FLA-Net~\cite{RN36}    & 2023\tiny{MICCAI} & full & 88.5 & -    & -    & -    & 87.4 & \textcolor{red}{88.0} & 85.6 & 85.8 & \textcolor{red}{85.7} \\
        MS-TFAL~\cite{RN37}    & 2023\tiny{MICCAI} & full & 91.1 & -    & -    & -    & 89.1 & \textcolor{red}{90.1} & 85.9 & 86.2 & \textcolor{red}{86.0} \\
        LGRNet~\cite{RN38}     & 2024\tiny{MICCAI} & full & 93.3 & -    & -    & -    & 91.6 & \textcolor{red}{92.5} & 87.5 & 87.6 & \textcolor{red}{87.5} \\
        EMCAD~\cite{RN41}      & 2024\tiny{CVPR}   & full & \textbf{95.2} & 92.3 & 92.3 & 92.8 & - & \textcolor{red}{92.6}   & - & - & -\\
        MedSAM2(Box)~\cite{RN56} & 2025  & full+prompt & 94.6 & 91.7 & \textbf{93.2} & 94.4 & 93.7 & \textcolor{red}{92.8} & - & - & -\\
        \midrule
        WeakPolyp~\cite{RN9}  & 2023\tiny{MICCAI} & weak & 85.2 & 74.5 & 71.1 & 86.1 & 84.8 & \textcolor{red}{76.7} & 79.2 & 80.7 & \textcolor{red}{79.8}\\
        \textbf{WeakPolyp+M2O} & - & weak & 89.2 & 75.9 & 72.4 & 90.0 & 90.0 & \textcolor{red}{79.0} & 81.1 & 81.0 & \textcolor{red}{81.0}\\
        \midrule
        SAM2(Point) & 2024 & full+prompt & 85.3 & 86.4 & 79.0 & 89.0 & 83.9 & \textcolor{red}{84.6} & 87.6 & 85.7 & \textcolor{red}{86.8} \\
        SAM2(Box) & 2024 & full+prompt & 93.0 & 92.3 & 91.0 & 89.7 & 92.4 & \textcolor{red}{91.7} & 93.1 & 92.6 & \textcolor{red}{92.9} \\
        SAM3(Point) & 2025 & full+prompt & 86.7 & 82.0 & 77.8 & 80.7 & 84.6 & \textcolor{red}{81.4} & 74.5 & 72.4 & \textcolor{red}{73.6} \\
        SAM3(Box) & 2025 & full+prompt & 93.4 & 92.4 & 91.5 & 89.9 & 94.6 & \textcolor{red}{92.1} & 94.2 & 93.4 & \textcolor{red}{93.9} \\
        \textbf{SAM2+OBBSeg(Point)} & \textbf{Ours} & weak+prompt & 91.9 & 87.3 & 86.9 & 91.2 & 92.1 & \textcolor{red}{88.5} & 90.3 & 89.7 & \textcolor{red}{90.0} \\
        \textbf{SAM2+OBBSeg(Scribble)} & \textbf{Ours} & weak+prompt & 92.4 & 89.2 & 90.4 & 94.2 & 92.6 & \textcolor{red}{90.6}  & 92.1 & 92.0 & \textcolor{red}{92.1} \\
        \textbf{SAM2+OBBSeg(Box)} & \textbf{Ours} & weak+prompt & 95.1 & 93.1 & 92.7 & 95.7 & 94.3 & \textcolor{red}{93.6} & {94.9} & 94.5 & \textcolor{red}{94.7}\\
        \textbf{SAM2+OBBSeg(Circle)} & \textbf{Ours} & weak+prompt & 95.1 & \textbf{93.8} & {92.9} & \textbf{95.7} & \textbf{94.8} & \textbf{\textcolor{red}{94.0}} & \textbf{95.1} & \textbf{94.8} & \textcolor{red}{\textbf{95.0}}\\
        \bottomrule
    \end{tabular}
    }
    \label{tab:table_1}
\end{table}

\begin{table}[htp]
    \centering
    \caption{Comparison of OBBSeg with fully supervised models on skin lesion, nodule and tumor datasets.}
    \resizebox{\textwidth}{!}{
    \begin{tabular}{l l c | c c c | c c c c}
        \toprule
        \multirow{2}{*}{\textbf{Methods}} & \multirow{2}{*}{\textbf{Year}} & \multirow{2}{*}{\textbf{Supervision}} & \multicolumn{3}{c|}{\textbf{Skin Lesion}} & \multicolumn{4}{c}{\textbf{Nodule \& Tumor}}\\
        \cmidrule(lr){4-6} \cmidrule(lr){7-10}
        & & & \textbf{ISIC2017} & \textbf{ISIC2018} & \textcolor{red}{\textbf{Avg.}} & \textbf{TN3K} & \textbf{DDTI} & \textbf{BUSI} & \textcolor{red}{\textbf{Avg.}}\\
        \midrule
        U-Net~\cite{RN6}     & 2015\tiny{MICCAI}  & full & 83.0 & 86.7 & \textcolor{red}{85.3} & 78.1 & 63.6 & 74.0 & \textcolor{red}{75.4}\\
        U-Net++~\cite{RN7}   & 2018\tiny{DLMIA}   & full & 83.0 & 87.5 & \textcolor{red}{85.8} & 63.8 & 65.1 & 74.8 & \textcolor{red}{65.6}\\
        AttnUNet~\cite{RN33}  &     -        & full & 83.7 & 87.1 & \textcolor{red}{85.8} & 79.9 & -    & 74.5 & \textcolor{red}{79.0}\\
        MEGANet~\cite{RN39}   & 2024\tiny{CVPR}    & full & 61.2 & 69.3 & \textcolor{red}{66.3} & 51.4 & 61.0 & 45.5 & \textcolor{red}{51.9}\\
        PraNet~\cite{RN8}    & 2020\tiny{MICCAI}  & full & 83.0 & 88.6 & \textcolor{red}{86.5} & 81.1 & 81.4 & 78.6 & \textcolor{red}{80.8}\\
        SANet~\cite{RN35}     & 2021\tiny{MICCAI}  & full & 84.8 & 88.1 & \textcolor{red}{86.9} & 83.8 & 83.6 & 80.1 & \textcolor{red}{83.2}\\
        TransUNet~\cite{RN34} &     -        & full & 85.0 & 89.1 & \textcolor{red}{87.6} & 81.8 & 71.5 & -    & \textcolor{red}{80.0}\\
        CASCADE~\cite{RN40}   & 2023\tiny{WACV}    & full & 85.5 & 90.4 & \textcolor{red}{88.6} & -    & -    & 79.2 & \textcolor{red}{79.2}\\
        EMCAD~\cite{RN41}     & 2024\tiny{CVPR}    & full & 86.0 & 91.0 & \textcolor{red}{89.1} & -    & -    & 80.3 & \textcolor{red}{80.3}\\        
        KnowSAM~\cite{RN42}   & 2025\tiny{TMI}     & semi & -    & 87.2 & \textcolor{red}{87.2} & 81.2 & 65.0 & -    & \textcolor{red}{78.4}\\
        \midrule
        WeakPolyp~\cite{RN9} & 2023\tiny{MICCAI}  & weak & 84.7 & 87.8 & \textcolor{red}{86.6} & 77.8 & 79.2 & 75.7 & \textcolor{red}{77.6}\\
        \textbf{WeakPolyp+M2O} & -  & weak & 85.2 & 89.3 & \textcolor{red}{87.8} & 78.8 & 80.0 & 77.7 & \textcolor{red}{78.8}\\
        \midrule
        SAM2(Point) & 2024  & full+prompt & 83.7 & 87.0 & \textcolor{red}{85.7} & 79.1 & 85.4 & 86.2 & \textcolor{red}{81.1} \\
        SAM2(Box) & 2024  & full+prompt & 87.1 & 90.2 & \textcolor{red}{89.0} & 81.3 & 90.2 & 89.5 & \textcolor{red}{83.8} \\
        SAM3(Point) & 2025 & full+prompt & 76.4 & 83.6 & \textcolor{red}{80.7} & 58.2 & 58.8 & 72.4 & \textcolor{red}{60.4}\\
        SAM3(Box) & 2025 & full+prompt & 90.2 & 95.6 & \textcolor{red}{93.5} & 87.4 & 88.9 & 88.6 & \textcolor{red}{87.8}\\
        \textbf{SAM2+OBBSeg(Point)}  & \textbf{Ours} & weak+prompt & 86.0 & 89.3 & \textcolor{red}{88.0} & 84.3 & 86.9 & 87.0 & \textcolor{red}{85.1} \\
        \textbf{SAM2+OBBSeg(Scribble)} & \textbf{Ours} & weak+prompt & 90.9 & 92.5 & \textcolor{red}{91.9} & 86.6 & 89.1 & 89.9 & \textcolor{red}{86.7} \\
        \textbf{SAM2+OBBSeg(Box)}      & \textbf{Ours} & weak+prompt & 93.1 & \textbf{95.0} & \textcolor{red}{94.3} & 88.8 & 93.3 & 92.7 & \textcolor{red}{89.0}\\
        \textbf{SAM2+OBBSeg(Circle)}   & \textbf{Ours} & weak+prompt & \textbf{93.7} & 94.8 & \textbf{\textcolor{red}{94.4}} & \textbf{91.2} & \textbf{94.3} & \textbf{93.2} & \textbf{\textcolor{red}{92.0}}\\
        \bottomrule
    \end{tabular}}
    \label{tab:table_2}
\end{table}

\begin{figure}[htp]
    \centering
    \includegraphics[width=1\textwidth]{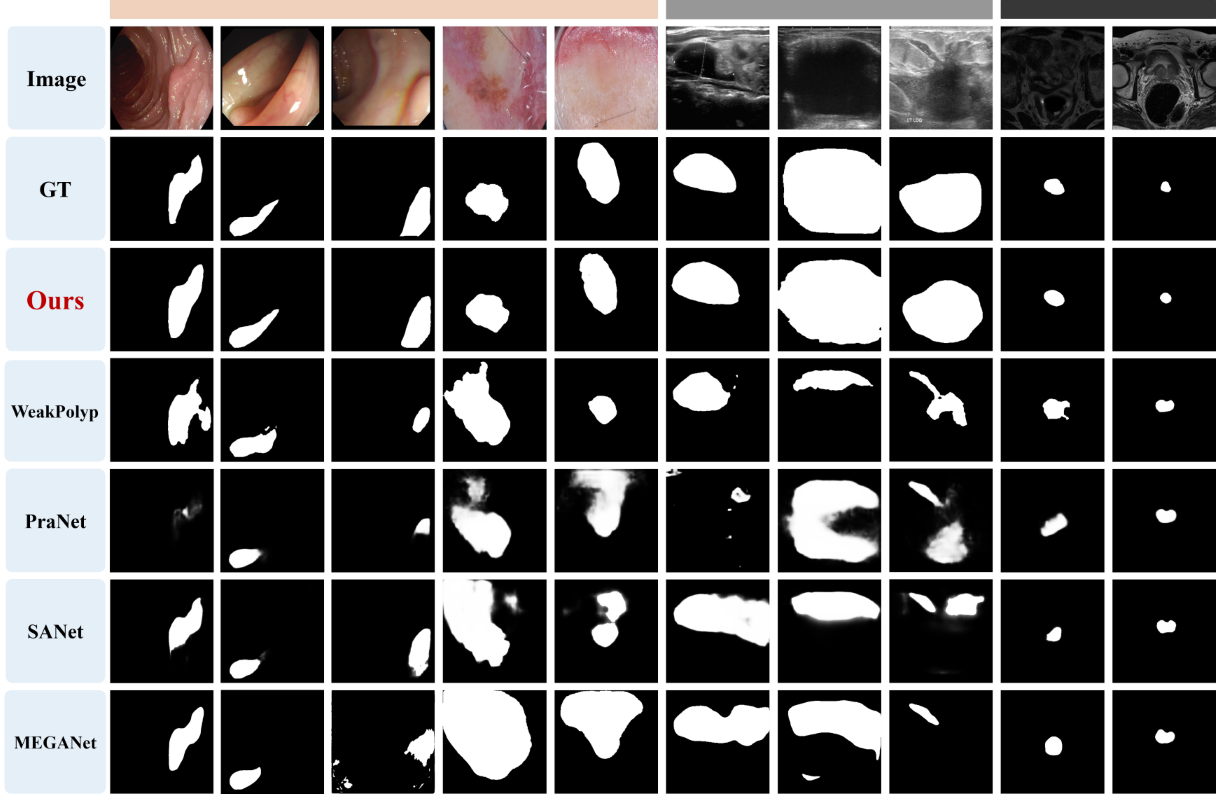}
    \caption{Visualization comparison across three modalities.}
    \label{fig:visualization}
\end{figure}

\textbf{Competing Methods.}
We compare OBBSeg with 20 representative methods, including U-Net~\cite{RN6}, U-Net++~\cite{RN7}, AttnUNet~\cite{RN33}, TransUNet~\cite{RN34}, PraNet~\cite{RN8}, FLA-Net~\cite{RN36}, MS-TFAL~\cite{RN37}, SANet~\cite{RN35}, LGRNet~\cite{RN38}, MEGANet~\cite{RN39}, CASCADE~\cite{RN40}, EMCAD~\cite{RN41}, WeakPolyp~\cite{RN9}, KnowSAM~\cite{RN42}, BoxInst~\cite{RN49}, BoxTeacher~\cite{RN50}, PointSup~\cite{RN51}, AGMM~\cite{RN52}, GazeMedSeg~\cite{RN53}, and MedSAM2~\cite{RN56}.

\textbf{Implementation Details.}
We use two backbones: Res2Net50~\cite{RN3} and the SAM2 encoder~\cite{RN12,RN31}. When using SAM2, we initialize the encoder with its pretrained weights and keep them frozen during training, and optimize only the OBBSeg-specific modules. Input images are resized to $352\times352$. Following WeakPolyp~\cite{RN9}, OBB annotations are generated from ground truth masks via random rotations and translations (Table~\ref{tab:obb_angle} and Table~\ref{tab:obb_relaxation}) to simulate real annotations. In the promptable setting, prompts are also simulated from ground truth masks following the standard protocol in promptable segmentation. These masks are used only offline to derive sparse or coarse prompt cues, and are not provided as dense training targets or loss supervision.

Training is performed on two RTX 4090 GPUs using AdamW (lr=$1e^{-4}$, weight decay=$5e^{-4}$) with a batch size of 16 for 16 epochs. For datasets without official splits, we adopt an 80/20 train–test split. During inference, OBBSeg can operate with prompts (SAM2+OBBSeg) or in a prompt-free manner, enabling both interactive and automatic segmentation.

\textbf{Evaluation Metrics.}
Model performance is evaluated using the Dice coefficient and the 95th percentile Hausdorff Distance (HD95). Dice measures overlap, while HD95 measures boundary accuracy. Higher Dice and lower HD95 indicate better performance. See the supplementary material for detailed definitions.

\begin{figure}[htp]
    \centering
    \begin{subfigure}{0.47\linewidth}
        \includegraphics[width=\linewidth]{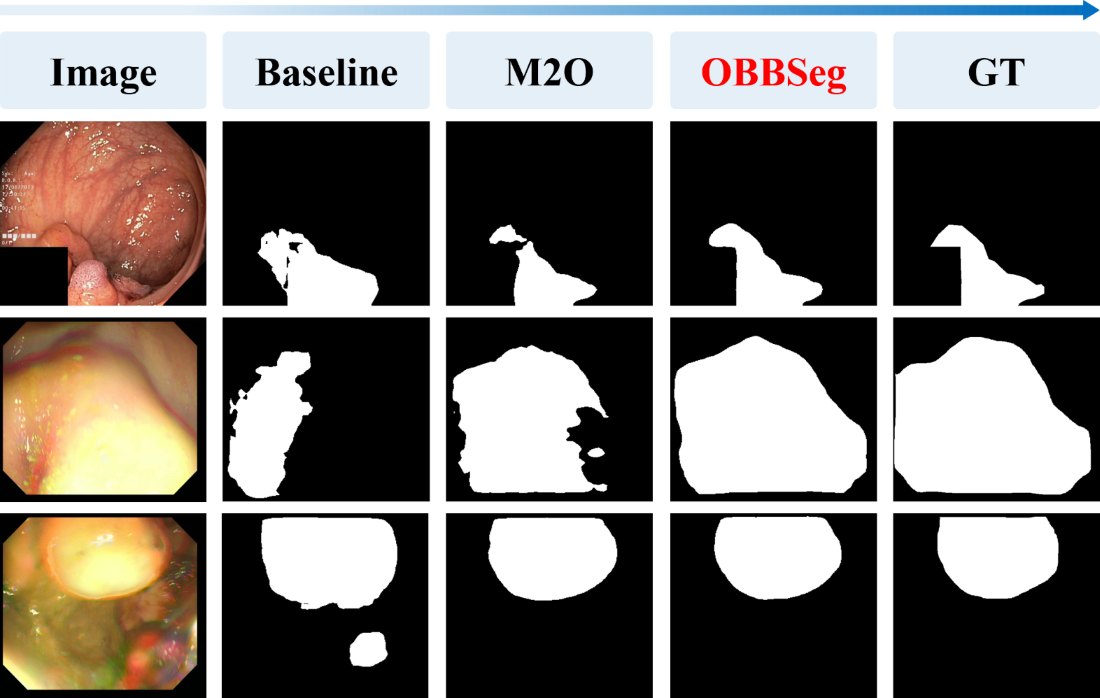}
        \caption{}
    \end{subfigure}
    \hfill
    \begin{subfigure}{0.5\linewidth}
        \includegraphics[width=\linewidth]{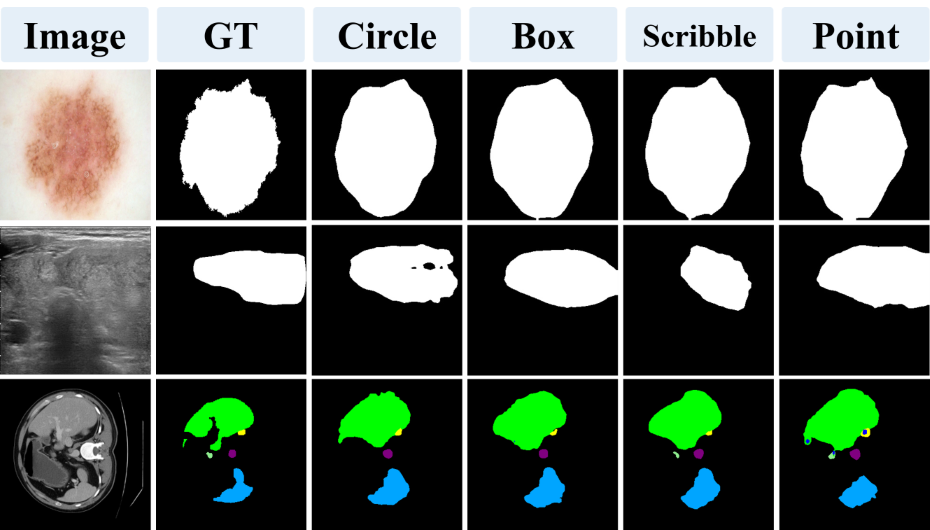}
        \caption{}
    \end{subfigure}
    \caption{(a) Impact of the proposed modules with the SAM2 backbone. (b) Visualization of representative failure cases and limitations.}
    \label{fig:visualization:compare}
\end{figure}

\begin{table}[htp]
    \centering
    \caption{Comparison with weakly supervised models.}
    \setlength\tabcolsep{20pt}
    \scriptsize
    \begin{tabular}{l | l | c c}
        \toprule
        {\textbf{Methods}} & \textbf{Supervision} & {\textbf{Kvasir}} & {\textbf{NCI-ISBI}}\\
        \midrule
        BoxInst~\cite{RN49} & Box &{65.7} & {73.78} \\
        AGMM~\cite{RN52} & Scribble & {67.2} & {72.7} \\
        PointSup~\cite{RN51} & Point & {73.0} & {73.4} \\
        BoxTeacher~\cite{RN50} & Box & {73.3} & {75.6} \\        
        AGMM~\cite{RN52} & Point & {75.5} & {73.8} \\
        GazeMedSeg~\cite{RN53} & Gaze & 77.8 & 77.6 \\
        \midrule
        \textbf{WeakPolyp+M2O} & OBB & 90.0 & 85.1 \\
        \textbf{OBBSeg(Ours)} & OBB+Point & 91.2 & 86.0 \\
        \textbf{OBBSeg(Ours)} & OBB+Scribble & 94.2 & 90.8 \\
        \textbf{OBBSeg(Ours)} & OBB+Box & 95.7 & 93.3 \\
        \textbf{OBBSeg(Ours)} & OBB+Circle & \textbf{95.7} & \textbf{94.1} \\
        \bottomrule
    \end{tabular}
    \label{tab:weakly_exp}
\end{table}

\begin{table}[htp]
    \centering
    \setlength\tabcolsep{9pt}
    \scriptsize
    \caption{Ablation study of $\mathcal{L}_{\text{M2O}}$, $\mathcal{L}_{\text{PS}}$, PAFE, and DBFE on five polyp datasets. "Base" denotes the baseline using the SAM2 encoder trained under OBB supervision, \textbf{rather than the original fully supervised SAM2}.}
    \begin{tabular}{c c c c c | c c c c}
        \toprule
        Base & {$\mathcal{L}_{\text{M2O}}$} & {$\mathcal{L}_{\text{PS}}$} & DBFE & PAFE & \multicolumn{4}{c}{without prompt}\\
        \midrule
        \Checkmark  &            &              &            &            & \multicolumn{4}{c}{77.2} \\
        \Checkmark  & \Checkmark &              &            &            & \multicolumn{4}{c}{80.5} \\
        \Checkmark  & \Checkmark &              & \Checkmark &            & \multicolumn{4}{c}{83.3} \\
        \Checkmark  & \Checkmark &              &            & \Checkmark & \multicolumn{4}{c}{81.3} \\
        \Checkmark  & \Checkmark &              & \Checkmark & \Checkmark & \multicolumn{4}{c}{\textbf{83.6}} \\
        \toprule
        \multirow{2}{*}{Base} & \multirow{2}{*}{$\mathcal{L}_{\text{M2O}}$} & \multirow{2}{*}{$\mathcal{L}_{\text{PS}}$} & \multirow{2}{*}{DBFE} & \multirow{2}{*}{PAFE} & \multicolumn{4}{|c}{with prompt} \\
        & & & & & {Circle} & {Box} & {Scribble} & {Point} \\
        \midrule
        \Checkmark  & \Checkmark & \Checkmark   &            &             & 82.5 & 82.4 & 78.5 & 78.2\\
        \Checkmark  & \Checkmark & \Checkmark   & \Checkmark &             & 91.8 & 91.5 & 86.4 & 81.7 \\
        \Checkmark  & \Checkmark & \Checkmark   & \Checkmark & \Checkmark  & \textbf{94.0} & \textbf{93.6} & \textbf{90.6} & \textbf{88.5} \\
        \bottomrule
    \end{tabular}
    \label{tab:ablation}
\end{table}

\begin{table}[htp]
    \centering
    \caption{Detailed segmentation results for all organs. Dice (in \%) is used for evaluation.}
    \setlength\tabcolsep{3pt}
    \scriptsize
    \begin{tabular}{lccccccccc}
    \toprule
    {\textbf{Prompts}} & Aorta & Gallbladder & Kidney(L) & Kidney(R) & Liver & Pancreas & Spleen & Stomach & \textcolor{red}{\textbf{Avg.}} \\
    \midrule
    Point & 85.8 & 86.3 & 84.6 & 85.0 & 86.3 & 80.4 & 82.9 & 79.4 & \textcolor{red}{83.8}\\
    Box &  90.5 & 81.3 & 82.6 & 91.4 & 85.5 & 85.2 & 87.6 & 88.2 & \textcolor{red}{86.5}\\
    Circle &  91.7 & 82.6 & 80.1 & 90.9 & 89.0 & 84.7 & 87.2 & 89.0 & \textcolor{red}{86.9}\\
    Scribble & 88.8 & 88.9 & 83.5 & 89.3 & 92.0 & 84.8 & 87.5 & 91.1 & \textcolor{red}{88.2}\\
    \bottomrule
    \end{tabular}
\label{tab:full_organ_results}
\end{table}

\textbf{Quantitative Comparison.}
Tables~\ref{tab:table_1} and~\ref{tab:table_2} report OBBSeg results on diverse segmentation tasks, using the weighted average Dice score (“Avg.”) based on dataset sizes.
Two baselines, WeakPolyp and SAM2, are considered. WeakPolyp uses Res2Net~\cite{RN3} as the backbone and does not require prompts. By simply adding the M2O loss without other modifications, we achieve a consistent \textbf{1\%–2\%} improvement across all datasets, demonstrating the effectiveness of M2O.
For SAM2, we integrate the full OBBSeg framework (M2O + Prompt) and evaluate different prompt types. This leads to substantial performance gains, approaching, and on several datasets exceeding, the performance of fully supervised methods across datasets.
These results demonstrate the effectiveness of combining weak supervision with prompt guidance, highlighting its potential for large-scale applications.
Table~\ref{tab:weakly_exp} further compares OBBSeg with existing weakly supervised methods. WeakPolyp+M2O already outperforms prior approaches by a large margin. Building upon this, OBBSeg with prompt guidance further improves performance, achieving state-of-the-art results on both the polyp (Kvasir) and prostate (NCI-ISBI) datasets.

\begin{figure}[htp]
    \centering
    \includegraphics[width=0.65\columnwidth]{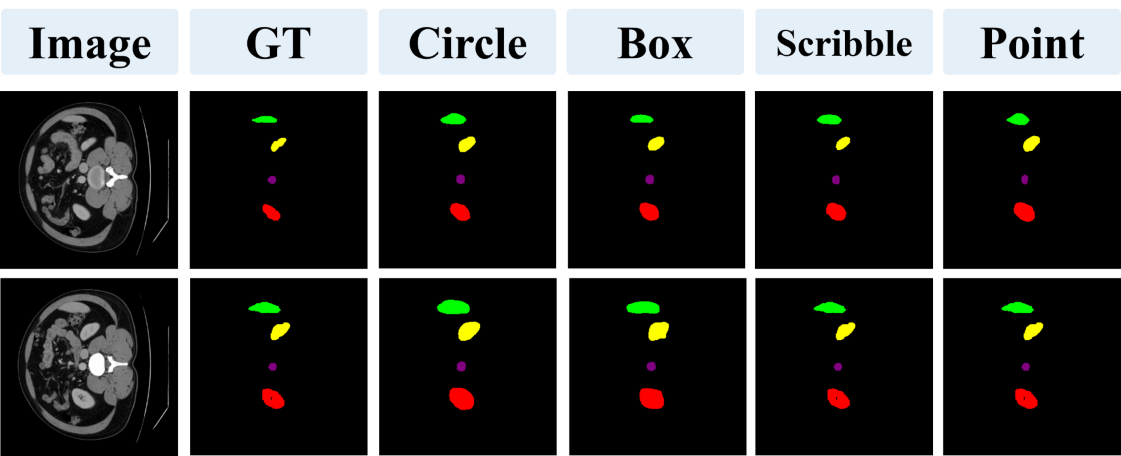}
    \caption{Visualization of multi-organ segmentation on Synapse dataset.}
    \label{fig:visualization_multi_class}
\end{figure}

\textbf{Multi-organ Segmentation. }
We evaluate OBBSeg on the 8-class Synapse dataset~\cite{RN55}, achieving average Dice of \textbf{83.8\%}, \textbf{86.5\%}, \textbf{86.9\%}, \textbf{88.2\%} across Point, Box, Circle, Scribble prompts in Table~\ref{tab:full_organ_results}, respectively, demonstrating its effectiveness for multi-class segmentation.

\textbf{Visualization Comparison.}
Fig.~\ref{fig:visualization} shows qualitative comparisons across three modalities. Despite using only coarse OBB supervision, our model (row 3) accurately captures lesion boundaries. Especially for elongated or tilted structures (column 1, 2, 3), OBBSeg not only outperforms the weakly supervised WeakPolyp but also achieves visually competitive results with fully supervised models(row 5, 6, 7), highlighting OBBSeg’s consistent superiority across diverse modalities.

\textbf{Ablation Study.}
Table~\ref{tab:ablation} and Fig.~\ref{fig:visualization:compare}(a) present the contribution of each module. M2O provides clear gains and establishes robust supervision with reduced box bias. When prompts are available, the prompt supervision loss yields the largest improvement, with circle prompts achieving the best performance.
DBFE and PAFE are complementary modules that benefit from accurate foreground cues provided by prompts. DBFE relies on foreground–background contrast estimation; however, noisy early predictions limit its effectiveness without prompt guidance. Prompt supervision supplies reliable foreground cues, enabling DBFE to better distinguish features and further improve performance.

\textbf{Boundary Segmentation Analysis.}
Table~\ref{tab:boundary:seg} reports boundary segmentation performance (HD95) under box and OBB annotations. OBB supervision achieves notably lower HD95 values, indicating more precise and smoother boundaries, confirming that OBBs provide stronger geometric constraints and better boundary localization.

\textbf{Performance under Equivalent Annotation Budget.}
To further validate the efficiency of Oriented Bounding Box (OBB) supervision, we conduct a comparative analysis under a fixed annotation budget in Table~\ref{tab:budget_comparison}. Based on the average annotation time per instance, we subsample the OBB training set such that its cumulative annotation cost aligns with that of the full axis-aligned bounding box (AABB) training set. 

\textbf{Elongated Lesion Segmentation.} 
Table~\ref{tab:elongated:seg} presents segmentation results for elongated lesions. Incorporating OBB and M2O loss improves performance, with particularly strong gains for slender lesions (+5.3\%), demonstrating that OBBs better capture the shape of elongated structures.

\begin{table}[htp]
    \centering
    \caption{Comparison of boundary segmentation using the HD95 index under box and OBB annotations (lower is better).}
    \setlength\tabcolsep{8.6pt}
    \scriptsize
    \begin{tabular}{l | l c c c c c | c c}
        \toprule
        & \multirow{2}{*}{\textbf{Methods}} & \multicolumn{5}{c|}{\textbf{Five Standard Polyp Datasets}} & \multicolumn{2}{c}{\textbf{SUN-SEG}} \\
        \cmidrule(lr){3-9}
        & & \textbf{Clinic} & \textbf{Colon} & \textbf{ETIS} & \textbf{Kvasir} & \textbf{Endo}  & \textbf{Easy} & \textbf{Hard}\\
        \midrule
        \multirow{4}{*}{\rotatebox[origin=c]{90}{Box}} & Point & 16.5 & 24.5 & 21.3 & 23.6 & 6.7 & 17.8 & 19.1 \\
        & Scribble & 13.1 & 17.0 & 10.1 & 15.9 & 8.6 & 15.1 & 15.7 \\
        & Box & 16.0 & 16.1 & 10.9 & 22.5 & 5.0 & 18.4 & 18.6 \\
        & Circle & 15.0 & 12.1 & 10.6 & 18.4 & 4.6 & 16.9 & 17.3 \\
        \midrule
        \multirow{4}{*}{\rotatebox[origin=c]{90}{OBB}} & Point & {9.2} & 16.6 & 13.0 & 16.7 & 5.7 & 12.0 & 12.6 \\
        & Scribble & {7.7} & 11.2 & 6.2 & 9.8 & 6.0 & 11.2 & 11.0 \\
        & Box & {4.7} & 5.9 & 5.8 & 6.8 & 4.3 & 7.1 & 7.0 \\
        & Circle & {5.2} & 5.4 & 4.5 & 8.2 & 3.5 & 6.9 & 6.6 \\
        \bottomrule
    \end{tabular}
    \label{tab:boundary:seg}
\end{table}

\begin{table}[htp]
    \centering
    \caption{Performance comparison under the same total annotation budget.}
    \setlength\tabcolsep{17.6pt}
    \scriptsize
    \label{tab:budget_comparison}
    \begin{tabular}{l|c c c c}
    \toprule
    Method & +Point & +Scribble & +Box & +Circle \\ 
    \hline
    Axis-aligned Box & 86.5\% & 89.2\% & 84.7\% & 93.2\% \\
    \textbf{OBB (Ours)} & \textbf{87.8\%} & \textbf{90.9\%} & \textbf{93.5\%} & \textbf{93.7\%} \\ 
    \bottomrule
    \end{tabular}
\end{table}

\begin{table}[htp]
    \centering
    \caption{M2O loss for elongated (aspect ratio $>2$) lesions}
    \setlength\tabcolsep{15.3pt}
    \scriptsize
    \begin{tabular}{l | c c | c c}
        \toprule
        \multirow{2}{*}{\textbf{Methods}} & \multicolumn{2}{c|}{\textbf{FSPD}} & \multicolumn{2}{c}{\textbf{SUN-SEG}}\\
        \cmidrule(lr){2-5}
        & {\textbf{Avg.}} & Elongated & {\textbf{Avg.}} & Elongated\\ 
        \midrule
        WeakPolyp~\cite{RN9} & {76.7} & {65.9} & {79.8} & {76.2}\\
        \textbf{WeakPolyp+M2O} & {79.0} & {71.2} & {81.0} & {81.5}\\
        \quad \quad -- & \textbf{(+2.3)} & \textbf{(+5.3)} & \textbf{(+1.2)} & \textbf{(+5.3)} \\
        \bottomrule
    \end{tabular}
    \label{tab:elongated:seg}
\end{table}

\begin{table}[htp]
    \centering
    \caption{OBB angle sensitivity. Max/Min values in each row are in red; $\Delta$Dice indicates the maximum performance drop.}
    \setlength\tabcolsep{10pt}
    \scriptsize
    \begin{tabular}{l | c c c c c c c | r}
        \toprule
        {Angle} & {0°} & {±5°} & {±10°} & {±15°} & {±20°} & {±25°} & {±30°} & {$\Delta$ Dice}\\
        \midrule
        Point & \textcolor{red}{88.5} & 88.4 & 87.8 & 87.6 & 87.5 & 87.6 & \textcolor{red}{86.6} & \textbf{-1.9}\\
        Scribble & \textcolor{red}{90.6} & 89.9 & 89.9 & 89.5 & 89.9 & 89.1 & \textcolor{red}{88.6} & \textbf{-2.0}\\
        Box & \textcolor{red}{93.6} & 93.2 & \textcolor{red}{93.1} & 93.2 & 93.2 & 93.2 & 93.2 & \textbf{-0.5}\\
        Circle & \textcolor{red}{94.0} & 93.2 & 93.0 & 92.8 & 92.3 & 92.6 & \textcolor{red}{91.7} & \textbf{-2.3}\\
        \bottomrule
    \end{tabular}
    \label{tab:obb_angle}
\end{table}

\begin{table}[htp]
    \centering
    \caption{Ablation study on OBB padding sensitivity; $\Delta$Dice indicates the performance drop.}
    \setlength\tabcolsep{13pt}
    \scriptsize
    \begin{tabular}{l | c c c c c | r}
        \toprule
        {Padding} & {0} & {0$\sim$5} & {0$\sim$10} & {0$\sim$15} & {0$\sim$20} & {$\Delta$ Dice}\\
        \midrule
        WeakPolyp~\cite{RN9} & \textcolor{red}{78.5} & 75.3 & 71.7 & 69.0 & \textcolor{red}{66.3} & \textbf{-12.2}\\
        \midrule
        Point & \textcolor{red}{88.5} & 87.1 & 87.1 & 84.3 & \textcolor{red}{81.3} & \textbf{-7.2}\\
        Scribble & \textcolor{red}{90.6} & 90.3 & 89.0 & 86.5 & \textcolor{red}{85.1} & \textbf{-5.5}\\
        Box & \textcolor{red}{93.6} & 92.9 & 91.5 & \textcolor{red}{88.3} & 88.4 & \textbf{-5.3}\\
        Circle & \textcolor{red}{94.0} & 92.6 & 91.4 & 87.8 & \textcolor{red}{87.5} & \textbf{-6.5}\\
        \bottomrule
    \end{tabular}
    \label{tab:obb_relaxation}
\end{table}

\begin{table}[htp]
    \centering
    \caption{Robustness of human OBB annotations under comparable deviations. Deviation is measured by $1-\mathrm{IoU}(B_{\text{ref}}, B_{\text{ann}})$ (\%); $\Delta$Dice indicates the performance drop.}
    \setlength\tabcolsep{16.5pt}
    \scriptsize
    \begin{tabular}{l | c c | c}
        \toprule
        \textbf{Method} & \textbf{Setting} & \textbf{Deviation} & {$\Delta$ Dice} \\
        \midrule
        WeakPolyp & Box (padding=$0{\sim}20$) & 19.74 & -12.2 \\
        WeakPolyp+M2O & Human OBB Annotation & 21.20 & -1.5 \\
        \bottomrule
    \end{tabular}
    \label{tab:human_obb_robustness}
\end{table}

\textbf{OBB Sensitivity Analysis.}
We evaluated the sensitivity of OBB annotation to rotational and positional errors. Table~\ref{tab:obb_angle} shows that varying OBB angle by ±30° causes only a minor performance drop ($<$2\%). Table~\ref{tab:obb_relaxation} shows that shifting the padding by up to 15 pixels has negligible effect, demonstrating strong tolerance to annotation errors.

We further evaluate the robustness of human OBB annotations introduced in Sec.~\ref{sec:intro}. As shown in Table~\ref{tab:human_obb_robustness}, human OBBs have a deviation of $21.20\%$, which is comparable to the strongest Box perturbation in Table~\ref{tab:obb_relaxation} (Padding $0{\sim}20$, deviation $19.74$). Under a similar perturbation level, WeakPolyp suffers a substantial drop of $12.2\%$ ($78.5 \rightarrow 66.3$), whereas WeakPolyp+M2O with human OBBs decreases by only $1.5\%$ ($79.0 \rightarrow 77.5 \pm 0.32$). This suggests that the proposed OBB supervision remains stable under realistic annotation variations.

\textbf{Model Size and Inference Time.}
OBBSeg can be integrated with different backbones, and the model size and inference time therefore depend on the selected architecture. Using SAM2 as the backbone, the model size is approximately 900 MB, with per-image inference times of 8.4 ms (Point), 12.3 ms (Scribble), 12.1 ms (Box), and 8.2 ms (Circle).


\textbf{Failure Cases and Limitations.}
Due to the coarse nature of weak supervision, the annotations lack fine-grained boundary details. In addition, the projection mechanism in the M2O loss provides limited supervision within the OBB region. As shown in Fig.~\ref{fig:visualization:compare}(b), the method may produce suboptimal segmentation for lesions with complex contours or indistinct boundaries.
Furthermore, the projection-based design may struggle with highly complex tree-like anatomical structures (\eg, vascular networks) where topology plays an important role. The current framework is also designed for static 2D images and does not explicitly model temporal dependencies in medical videos. Extending the framework to native 3D medical volumes remains an important direction for future work.

\section{Conclusion}
We introduced OBBSeg, a new supervision paradigm that combines oriented bounding boxes with prompt-guided learning to provide stronger geometric constraints than conventional weak labels. The proposed Mask-to-OBB loss and hierarchical prompt supervision alleviate box-induced bias and significantly improve segmentation performance. Extensive experiments show that OBBSeg achieves accuracy approaching fully supervised methods across diverse datasets.
Nevertheless, segmentation quality is still limited by the coarse nature of OBB annotations. Future work will explore adaptive label refinement and hybrid supervision to further enhance scalability and generalization.

\section*{Acknowledgments}
This work was supported in part by ICFCRT (W2441020), Guangdong Basic and Applied Basic Research Foundation (2023B1515120026), Shenzhen Peacock Program (KQTD20210811090044003), Young Faculty Startup Fund of Shenzhen University, and Scientific Development Fund from Guangdong Provincial Key Laboratory of Visual Media and Multidimensional Intelligence.

\bibliographystyle{splncs04}
\bibliography{main}
\end{document}